\documentclass[10pt,twocolumn,letterpaper]{article}

\usepackage[pagenumbers]{cvpr} 

\newcommand{\mb}{\mathbf}

\definecolor{cvprblue}{rgb}{0.21,0.49,0.74}
\usepackage[pagebackref,breaklinks,colorlinks,allcolors=cvprblue]{hyperref}

\def\paperID{16812} 
\def\confName{CVPR}
\def\confYear{2025}

\title{Uncertainty Quantification in Continual Open-World Learning}

\author{Amanda Rios\\
Intel\\
{\tt\small amanda.rios@intel.com}
\and
Ibrahima Ndiour\\
Intel\\
{\tt\small ibrahima.j.ndiour@intel.com}
\and
Parual Datta\\
Intel\\
{\tt\small parual.datta@intel.com}
\and
Jerry Sydir\\
Intel\\
{\tt\small jerry.sydir@intel.com}
\and
Omesh Tickoo\\
Intel\\
{\tt\small omesh.tickoo@intel.com}
\and
Nilesh Ahuja\\
Intel\\
{\tt\small nilesh.ahuja@intel.com}
}

\begin{document}
\maketitle
\vspace{-3mm}
\begin{abstract}
AI deployed in the real-world should be capable of autonomously adapting to novelties encountered after deployment. Yet, in the field of continual learning, the reliance on novelty and labeling oracles is commonplace albeit unrealistic. This paper addresses a challenging and under-explored problem: a deployed AI agent that continuously encounters unlabeled data - which may include both unseen samples of known classes and samples from novel (unknown) classes - and must adapt to it continuously. To tackle this challenge, we propose our method COUQ "Continual Open-world Uncertainty Quantification", an iterative uncertainty estimation algorithm tailored for learning in generalized continual open-world multi-class settings. We rigorously apply and evaluate COUQ on key sub-tasks in the Continual Open-World: continual novelty detection, uncertainty guided active learning, and uncertainty guided pseudo-labeling for semi-supervised CL. We demonstrate the effectiveness of our method across multiple datasets, ablations, backbones and performance superior to state-of-the-art. We will release our code upon acceptance.
\end{abstract}   
\footnote[1]{Manuscript Under Review}
\vspace{-3mm}
\section{Introduction}
\label{sec:intro}
\vspace{-1mm}
Real-world AI systems frequently face evolving data distributions due to changes in operating conditions and the emergence of new classes after deployment. To ensure a robust response, AI systems should ideally be able to detect these novelties and continuously learn from them, while minimizing computing and labeling costs.
Early research on continual learning (CL) focused on the important problem of catastrophic forgetting \cite{french1999catastrophic,parisi2019continual} but relied on a so-called `Oracle' for two critical functions: (i) identifying novel test samples and (ii) providing labels for these samples. While beneficial for advancing initial CL research, the assumption of an omniscient oracle is unrealistic for real-world applications. In practice, adaptive AI systems should be capable of automatically identifying novelties, a challenging task known as novelty detection or out-of-distribution (OOD) detection. Further, learning from novelties usually requires labels for the novel samples (e.g. open-world classification \cite{parmar2023open}); yet, annotation of all such samples is expensive and impractical. 
Therefore, for data-efficient model updates, it is advantageous to label only a a small subset of informative data samples judiciously chosen from the larger pool of novel samples; a task known as sample selection in active learning \cite{lewis_gale,tong_koller,brinker,ren_survey_2021,settles2009active}.

\begin{figure}
\centering
\includegraphics[width=\linewidth]{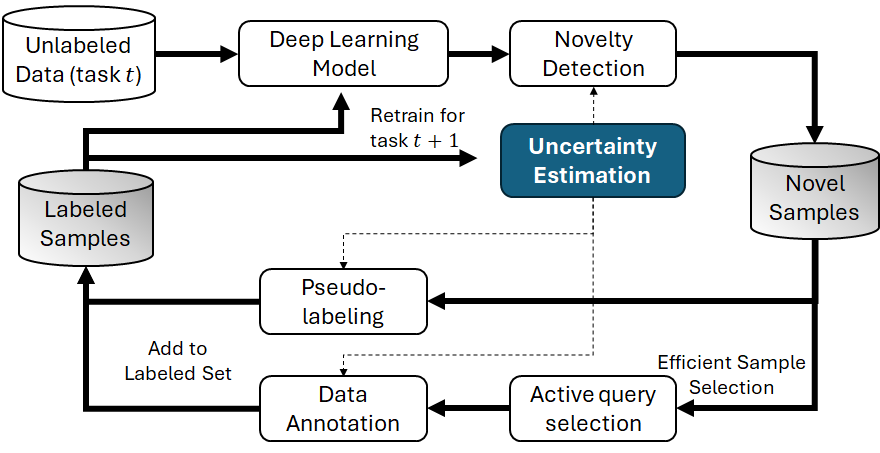}
\vspace{-5mm}
\caption{A general open-world, continual learning pipeline. Uncertainty estimation plays a crucial role in various points.}
\label{fig:CL_pipeline}
\vspace{-5mm}
\end{figure}

Although both these problems of novelty-detection and active sample-selection have been widely explored, most research has been in non-continual settings. 
There is extensive research on cost-effective active labeling \cite{yoo2019learning, sener2018active, gal2017deep} for conventional non-continual training, but surprisingly this has found little adoption in CL models \cite{Aguilar_2023_ICCV,vu2023active,ayub2022few}. Moreover, solutions for unsupervised and semi-supervised CL are scarce \cite{ordisco,Boschini_2022,bagus2022supervised} and often operate with a very strong oracle assumption: that in unlabeled input, past classes do not appear in conjunction with newly introduced classes - bypassing the need for novelty detection. For such methods, removing the oracle causes significant performance degradation. Methods under Generalized Category Discovery (GCD) \cite{vaze2022gcd,ma2024active} do address the scenario in which both old and new classes co-occur in the test data. 
However, they are designed for single-task setups where the entire unlabeled dataset is presented to the model at once for categorization, rather than in a continual, class-incremental manner. These solutions are ill-suited for (post-deployment) continual learning as they assume full data availability (labeled and unlabeled) and perform expensive full re-trainings poorly scalable to CL. Similarly, most solutions for novelty or out-of-distribution (OOD) detection \cite{hendrycks2016baseline, liang2018enhancing, lee2018simple, haoqi2022vim, ren2019likelihood} were developed for and evaluated against a single fixed binary partition of known (old) versus novel classes and not on continual splits. Such conventional OOD models are not designed to continually integrate and learn from the detected novel data. 

Overall, both novelty detection and active learning depend on the availability of reliable, high-quality uncertainty estimates, see figure \ref{fig:CL_pipeline}. A variety of uncertainty quantification techniques have been studied: softmax probability and its temperature-scaled variants \cite{hendrycks2016baseline,liang2018enhancing}; predictive uncertainty from Bayesian Neural Networks \cite{gal2016dropout}; feature-based Mahalanobis distances \cite{lee2018simple} and its variants \cite{ahuja2019bdl_dfm}. These methods work well in fixed settings where the model is trained once. However, in CL settings where the model parameters are continually updated,
the quality of uncertainty estimates deteriorates as knowledge from novel tasks and classes is incrementally integrated. Such estimates, if used either for novelty detection or for active sample selection, will lead to poor prediction of novelties and/or a poor choice of samples selected for CL. Furthermore, errors tend to accumulate as new tasks are encountered, resulting in progressively poorer performance over time. A recent model, incDFM \cite{rios2022incdfm}, attempted to address this problem by proposing a solution to continual novelty detection and integrated it into the broader pipeline of unsupervised class incremental learning. However, the number of novel classes introduced at each task was limited to one, so that any detected novelty could then be trivially labeled as belonging to a single novel class. The more general scenario of class increment learning with multiple novel classes was not addressed, which introduces significant new challenges: the number of new classes is not known \textit{a priori}, and  error-propagation is exacerbated since a novel class sample may not only be misidentified as an old-class sample (the only type of wrong prediction possible in incDFM), but could also be labeled incorrectly.

\vspace{-5mm}
\paragraph{Contribution:} We present an iterative uncertainty estimation technique suited for learning in an open-world, continual multi-class incremental environment wherein, at each continual task, a model is exposed to a mix of data from both known and an arbitrary number of unknown classes. To the best of our knowledge, this is the first work to address uncertainty estimation in such a general scenario. We show that unlike traditional one-shot uncertainty estimation methods, the quality of our uncertainty estimates \textit{does not} degrade as new classes are iteratively encountered and learned. We demonstrate its effectiveness by applying the technique to various tasks in a continual learning pipeline such as (i) continuous novelty detection in unsupervised and semi-supervised continual settings, and (ii) sample selection for active learning and for pseudo-labeling in semi-supervised continual learning. One advantage of our formulation is that it enables us to distinguish between \emph{confidently novel} samples (which are confidently identified as novel) as opposed to \emph{ambiguously novel} samples. This allows us to experiment with and analyze the effectiveness of various active sample selection strategies. Through exhaustive experiments and ablations, we demonstrate the superiority of our method for both these tasks across multiple classification datasets and various DNN backbones.
Results are benchmarked against multiple competitive baseline methods. 
\section{Problem Statement - Background}
\label{sec:problem}
\textbf{2.1. Problem Setting:} Consider a deep neural network model $y=\Phi(x)$ expected to learn from a sequence of continual tasks. $\Phi$ can be partitioned into a backbone network that produces features $u=g(x)$ followed by a classifier $y=f(u)$ operating on those features to produce the final prediction $y$, i.e. $\Phi = f \circ g(x)$. At each continual task,  $t$, the model is presented with an initially unlabeled set of samples $U(t)$ comprising a mixture of unseen samples of ``old''/learned classes  $U_{old}(t)$ and unseen samples from new/unlearned classes $U_{new}(t)$, 
\begin{align*}
U(t) &= U_{old}(t) \cup U_{new}(t), \text{where} \\
U_{old}(t)  &= \{X | X \thicksim \bigcup_{k=1}^{t-1} D_k\}, U_{new}(t) = \{X | X \thicksim D_t\},
\end{align*}
where $D_t$ comprises data from the set of new classes $C^t_{new} = \{c_i^t \}, i=1, \dots, N_t $, introduced at task $t$. For notational convenience, we denote $C_{old}^{t} = \bigcup_{k=0}^{t-1}C_{new}^k$, which is the collection of classes observed up to and including task $t-1$. Note that samples in $U_{old}(t)$ belong to previous classes but are ``unseen'', i.e., were never used in training during prior tasks. The goal is to accurately differentiate between $U_{old}(t)$ and $U_{new}(t)$ and simultaneously learn to process/classify the novel classes present in $U_{new}(t)$. This implies gaining the ability to classify/distinguish among different novel classes in $U_{new}(t)$. To accomplish this, we propose a multi-class continual uncertainty quantification algorithm that is able to generate uncertainty scores per detected novel class. The uncertainty scores can also be used for (i) selecting informative novel samples for active labeling, and (ii) selecting confidently novel samples for unsupervised pseudo-labeling.

\vspace{-3mm}
\paragraph{2.2. Other Relevant Solutions:}
The described problem setting is complex and can be broken down into a series of sub-problems (and sub-solutions) that will be outlined below. 
Some of the (sub-)solutions will be used as comparison methods to benchmark our approach in \S\ref{sec:exp_and_results}.
\vspace{-2mm}
\paragraph{2.2.1. Learning from Novelties:}
Early works on ``Open World Classification'' \cite{fei2016breaking, shu2017doc} were essentially the same as OOD detection and did not 
focus on learning from the detected novel data. More recent works on ``Category Discovery'' attempt to do this \cite{han2019learning} by estimating the number of novel classes and assign each novel sample to the appropriate novel class. They assume, however, that the unlabeled set contains only novel-class data ($U=U_{new}$), thus not requiring novelty detection. 
Work on ``Generalized Category Discovery'' (GCD) \cite{vaze2022gcd} removes this restriction and permit $U$ to include both known and novel classes. 
While this represents a significant advancement, it remains a challenging problem that, to the best of our knowledge, has only been addressed in non-continual settings \cite{vaze2022gcd,ma2024active}, without the added complexities inherent to continual learning.

\vspace{-3mm}
\paragraph{2.2.2. Continual Learning:} Most CL approaches that focus on mitigating catastrophic forgetting are fully supervised and assume access to fully labeled data streams \cite{rios2020lifelong,parisi2019continual,kirkpatrick2017overcoming,rios2018closed,rebuffi2017icarl,wen2021beneficial,cheung2019superposition}. More recent approaches, though, do not make this assumption and explore unsupervised, semi-supervised, and few-shot continual learning methods \cite{ordisco,Boschini_2022,bagus2022supervised}. 
However, these too assume that the incoming data contains only novel classes ($U(t) = U_{new}(t)$), thus bypassing the need for novelty detection and avoiding error propagation. In this category, we compare to CCIC \cite{Boschini2022}, a semi-supervised CL method that leverages the MixMatch technique \cite{berthelot2019mixmatch} to learn more efficiently from both labeled and unlabeled samples. As shown in \S\ref{sec:exp_and_results}, approaches like CCIC scale poorly to our generalized setting where the incoming data can include both old and novel classes.
Lastly, continual novelty detection (CND) remains under-explored \cite{aljundi2022continual}, with a notable exception of incDFM \cite{rios2022incdfm}, which constrained $U_{new}(t)$ to have only one new class at a time. We compare our approach to incDFM in the results \S\ref{sec:exp_and_results} and show that it does not generalize well when $U_{new}(t)$ may contain an arbitrary number of novel classes.

\vspace{-3mm}
\paragraph{2.2.3. Active Learning:} Active learning aims to learn from a small set of informative data samples judiciously chosen from a larger unlabeled dataset. Diverse strategies have been used for selecting the samples based on uncertainty \cite{lewis_gale,tong_koller} or diversity \cite{brinker}. We refer to \cite{ren_survey_2021,settles2009active} for exhaustive surveys of the methods, which overwhelmingly operate in an offline fashion. Defining an effective AL heuristic in the generalized setting of $U(t)$ is challenging \cite{vu2023active} as will be shown in \ref{sec:exp_and_results}.
An early attempt was made by the authors of GBCL \cite{GBCL}. Their method integrates AL with few-shot continual learning by selecting samples that are most distant from a continually updated Gaussian mixture model of all old classes. We compare to GBCL in \S\ref{sec:CONL}.

\section{Our Approach: COUQ}
\label{sec:method}

\paragraph{Choice of elemental uncertainty metric:} We refer to our method as COntinual Uncertainty Quantification (COUQ). Our method is agnostic to the choice of the underlying uncertainty measure used within our algorithm, as long as it can reliably estimate uncertainty per novel class or per old class. However, this is not an easy feat since many existing static novelty detection approaches make for very poor per-class uncertainty estimators. In our current formulation, we leverage the \textit{feature reconstruction error} (FRE) metric introduced in \cite{ndiour2020probabilistic}, which has been shown to effectively estimate per-class uncertainty in the non-continual setting. For each in-distribution class, FRE learns a PCA (principal component analysis) transform $\{\mathcal{T}_m\}$ that maps high-dimensional features $u$ from a pre-trained deep-neural-network backbone $g(x)$ onto lower-dimensional subspaces. During inference, a test-feature $u=g(x)$ is first transformed into a lower-dimensional subspace by applying $\mathcal{T}_m$ and then re-projected back into the original higher dimensional space via the inverse $\mathcal{T}_m^{\dagger}$. 
The FRE measure is calculated as the  $\ell_2$ norm of the difference between the original and reconstructed vectors:
\vspace{-2mm}
\begin{equation}
\label{eq:FRE}
    FRE_m(u) = \|f(\mb{x})-(\mathcal{T}_m^{\dagger} \circ \mathcal{T}_m)u\|_2.
    \vspace{-2mm}
\end{equation}
Intuitively, $FRE_m$ measures the distance of a test-feature to the distribution of features from class $m$. If a sample does not belong to the same distribution as that $m$th class, it will usually result in a large reconstruction score $FRE_m$. FRE is particularly well suited for continual settings since for each new class an additional PCA transform can be trained without disturbing the ones learnt for previous classes.

\vspace{-1mm}
\subsection{Algorithmic Steps}
\label{sec:steps-approach}
\paragraph{Initial training and deployment:} At the outset (i.e. task $t=0$), we assume that the main model $\Phi(x)$ has been trained to classify among an initial fixed set of classes $C_{new}^0$ (following notation from Sec.\ref{sec:problem}). An initial set of PCA transforms $\{\mathcal{T}^0_m\}, m\in C_{new}^0$ have also been learnt.

\vspace{-5mm}
\paragraph{Continual Learning and Adaption:} At any task $t, t>0$, as unlabeled data arrives, COUQ will follow an iterative procedure to derive uncertainty scores that can be used to detect novelties and classify them if present. Classification of novelties can be performed in an unsupervised manner using clustering techniques such as K-means, or in a semi-supervised manner via a combination of active and Pseudo labeling. We do not prescribe a particular approach; rather, we show that our uncertainty estimation algorithm can work well with both semi-supervised and unsupervised approaches. In the former case, the selection of samples for active labeling itself can be guided by the uncertainty score from COUQ. Finally, note that this iterative procedure is an inner-loop iteration (indexed by $i$) employed at each task, which is different from the outer-loop iteration over tasks (indexed by $t$). Each iteration $i$ utilizes the scores and novelty class predictions from the previous iteration to progressively obtain better uncertainty scores $S^{t,i}(u)$. To simplify the notation, we index only w.r.t iteration $i$, with the understanding that uncertainties are recalculated at every task.

\vspace{-4mm}
\paragraph{$i=0$: Initializing COUQ.} At the first inner-loop iteration, uncertainty scores $S^0(u)$ are simply 
\begin{equation}
\label{eq:uncertainty_scores}
S^0(u) = \min_{j\in C_{old}^t}FRE_j^0(u),
\vspace{-2mm}
\end{equation}
\noindent
$FRE^0$ indicates scores at the $0^\text{th}$ iteration of the $t^\text{th}$ task. These reflect the distance to the detected past classes encountered till $t-1$. 
This score can be used for novelty detection by noting that samples with high $S^0(u)$ values have a significant probability of being novel. Once confidently novel samples have been identified, these require a `novelty mapper' $M^i(u)$ to assign labels or IDs for the subsequent iterations. This mapper can be obtained from an unsupervised approach such as K-means clustering by selecting the id of the closest cluster centroid. 
Alternately, we can choose a small number $b_0$ samples with high $S^0(u)$ values for active querying and train a pseudo-labeler (such as a small MLP) on these $b_0$ samples to predict the class-ids for the remaining confidently novel samples. Either way, the novel classes that are identified 
are appended to $C_{new}^t$ and are used for computing initial estimates of per-class PCA transforms for the new classes $\{\mathcal{T}^{t,i=0}_m\}, m\in C_{new}^t$. 

\vspace{-2mm}
\paragraph{$i>0$: Iterative Training.} For all subsequent iterations of the same task, COUQ  computes a per-novel-class uncertainty score, relying on the previous iteration's mapper novel class-id predictions and the corresponding previous iterations per-novel-class PCA transforms $\{\mathcal{T}^{t,i-1}_m\}$. The overall multiclass uncertainty score for a given unlabeled sample $u$ is defined in eq \ref{eq:uncertainty-uncanny}. A novel class-id pseudo-label $m$ is predicted by the novelty mapper from the previous iteration $M^{i-1}(u)$, and we select the corresponding PCA transform $\mathcal{T}^{t,i-1}_m$ to calculate the score. 
\begin{equation}
    S^i(u) = \min_{j\in C_{old}^t}\frac{FRE_j^0(u)}{FRE^{i-1}_m(u)}; 
    m = M^{i-1,t}(u) \in C^t_{new}
\label{eq:uncertainty-uncanny}
\end{equation}

\noindent
This score can be used to broadly categorize samples in $U(t)$ as follows:
\begin{enumerate}
    \item \emph{Identify as novel with high-confidence}: These are samples with the highest score values, which will occur for a high numerator relative to the denominator. A high value of numerator implies large distance from previously seen classes $C_{old}^t$, while a low value of the denominator implies low distance from novel class $m$. Such a sample likely belongs to $U_{new}(t)$ and is a strong candidate to be pseudo-labeled as class $m$.

    \item \emph{Identify as old-class with high-confidence}: This is the opposite of the previous case: low score values corresponding to low numerator (low distance w.r.t $C_{old}^{t-1}$) and high numerator value (high-distance from the novel class $m$). Such a sample likely belongs to $U_{old}(t)$ and is not needed any further for novelty detection (since we are assuming no distribution-shift for old classes).

    \item \emph{Ambiguous}: Samples for which the score is neither definitively high or low. These could be old-class samples having relatively high scores, or novel-class samples having relatively low scores. Owing to this ambiguity, a clear determination cannot be made, and hence these samples would most benefit from active labeling.  
\end{enumerate}

We iteratively improve the quality of our multiclass uncertainty measure, $S^i(u)$, via a pseudo-labeling and/or active-labeling at each iteration. Using the novel class-id predictions from iteration $i-1$ (un-supervised or semi-supervised), we separate and sort the scores per predicted novel class $m \in C_{new}^t$. We select the topmost $\alpha$ percent per predicted-class. These are the samples predicted as novel class $m$ with highest-confidence. If active querying is also used, then we additionally select $b_i$ most ambiguous samples per predicted novel class to actively label, so long as the tiny active label budget has not been exhausted. 
Together, the accumulated selected active or pseudo-labeled samples are used for (i) computing all novel PCA transformations $\{\mathcal{T}^{t,i}_m\}$, (ii) re-training the novelty mapper - either the unsupervised K-means or the small MLP pseudo-labeler - in preparation for the next iteration, and (iii) updating the main model $\Phi$ classifier (or only the classifier $f$ if using a frozen backbone $g$).
Further details on COUQ iterations, such as stopping criteria, as well as a detailed Algorithm flow box are included in supplementary.

\subsection{Application to Continual Novelty Detection}
\vspace{-1mm}
As described, uncertainty estimation in COUQ can be used directly for novelty detection. Importantly, because the addition of new PCA transforms (per detected novel class) does not impact those already stored in memory, novelty detection performance does not significantly degrade, e.g. does not "catastrophically forget". Furthermore, the use of an iterative approach results in higher-quality and more consistent uncertainty estimates than other one-shot approaches, thus also minimizing continual error propagation.

\vspace{-1mm}
\subsection{Applications to Active Sample Selection} 
\label{sec:COUQ-AL}
\vspace{-1mm}
We describe next how the uncertainty score from Eq. (\ref{eq:uncertainty_scores}) can be used for efficient active annotation. The samples thus labeled can be used not only for improving the quality of the multiclass uncertainty measure as desribed above, but also for updating the weights of a downstream continual classifier $f$. 
One possible sample selection strategy would be to prioritize novel samples for annotation. This could be done by selecting the most confident novel samples for active annotation. However, we find that including samples of ambiguous novelty, scores $S^i(u)$ which are neither too high or too low, is more informative. We set our default AL strategy to pick 1:1 between ambiguous and confident novel samples per detected novel class. Details of how ambiguous samples are determined can be found in the supplementary. In \S\ref{sec:CONL-AL}, we compare various sample selection strategies using different uncertainty scores and show that our proposed AL strategy using $S^i(u)$ uncertainty scores significantly outperform others.

\label{sec:exp_and_results}
\begin{figure*}[t]
\centering
\includegraphics[width=17cm]{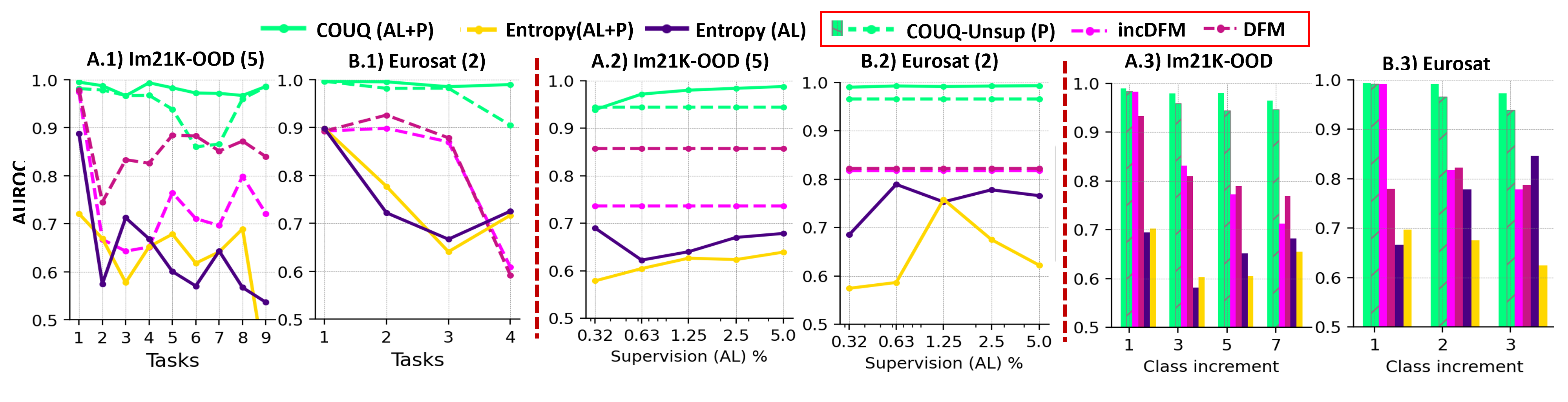}
\vspace{-5mm}
\caption{(\textit{Left} A.1,B.1) AUROC of Novelty Detection at each continual task. Number of novel classes per task is in parenthesis. COUQ (green) clearly outperforms baselines both in both semi-supervised (solid line) and unsupervised versions (dashedline); (\textit{Center} A.2,B.2) Results varying the supervision budget; (\textit{Right} A.3,B.3) Results varying Novel Class Increment per task. For (left,right) Supervision budget is 1.25\% and all plots show results implemented with a Resnet50 backbone. Equivalent plots for other datasets in appendix.}
\vspace{-2mm}
\label{fig:plots_CND}
\end{figure*}

\begin{table*}[t]
\scriptsize
\centering
\addtolength{\tabcolsep}{-0.23em}
\begin{tabular}
{c|ccc|ccc|ccc|ccc|ccc|c} 
\toprule
 & \multicolumn{3}{c|}{Im21K} & \multicolumn{3}{c|}{Eurosat} & \multicolumn{3}{c|}{Plants} & \multicolumn{3}{c|}{Cifar100} & \multicolumn{3}{c}{Places} & Average \\ 
Method & R50 & ViT-B & ViT-S & R50 & ViT-B & ViT-S & R50 & ViT-B & ViT-S  & R50 & ViT-B & ViT-S & R50 & ViT-B & ViT-S & All\\ 
\midrule 
COUQ (AL+P; ours) & \textbf{98.0} & \textbf{94.5} & \textbf{93.5} & \textbf{99.2} & \textbf{99.0} & \textbf{97.2} & \textbf{83.4} & \textbf{69.9} & \textbf{67.2} & \textbf{81.6} & \textbf{87.0} & \textbf{83.3} & \textbf{78.7} & \textbf{69.3} & \textbf{69.2} & \textbf{84.7}\\ 
ER-Entropy & 65.0& 60.1& 56.7& 77.4& 62.2& 61.4& 58.2& 53.4& 55.6& 61.2& 56.4& 41.6& 57.5& 52.3& 54.7& 58.2\\ 
ER-Margin & 66.5& 54.8& 57.2& 69.0& 49.8& 60.4& 58.7& 54.6& 52.0& 60.6& 54.5& 52.1& 58.5& 53.0& 53.4& 57.0\\ 
ER-Softmax & 64.9& 54.0& 58.1& 76.7& 52.5& 59.0& 57.2& 54.7& 53.9& 61.2& 50.1& 53.7& 58.9& 55.3& 54.2& 57.6\\ 
PseudoER-Entropy & 58.9& 53.3& 54.4& 62.5& 52.9& 51.1& 58.1& 57.2& 52.0& 62.2& 53.4& 52.3& 53.1& 51.7& 53.4&55.1\\ 
PseudoER-Margin & 60.9& 53.7& 52.8& 56.7& 51.9& 50.7& 60.8& 53.7& 57.0& 61.0& 53.5& 52.3& 54.6& 53.1& 51.0& 54.9\\
PseudoER-Softmax & 62.4& 53.0& 51.4& 47.0& 49.2& 50.6& 58.9& 54.5& 53.7& 61.6& 54.3& 53.6& 56.1& 52.1& 50.4& 53.9\\
\midrule 
COUQ-Unsup (P; ours) & \textbf{93.9} & \textbf{86.7} & \textbf{84.0} & \textbf{96.7} & 86.3 & \textbf{80.8} & \textbf{81.3} & 64.0 & \textbf{61.1} & \textbf{81.5} & \textbf{81.5} & \textbf{80.9} & \textbf{77.4} & \textbf{63.4} & \textbf{66.6} & \textbf{79.1}\\ 
incDFM & 77.3& 69.1& 76.0& 81.8& \textbf{91.9}& 74.9& 68.7& 63.2& 58.2& 74.6& 68.0& 66.3& 68.6& 63.0&64.7 & 71.1\\ 
DFM & 79.0& 85.7& 75.4& 82.2& 80.7& 74.9& 67.4& \textbf{65.2}& 54.9& 73.5& 68.9& 66.3& 68.4& 62.9& 64.1& 71.3\\ 
\midrule 
\end{tabular}
\vspace{-3mm}
\caption{Continual Novelty Detection measured by AUROC varying backbone implementation. Supervision budget is 1.25\% with the exception of the unsupervised methods, e.g. COUQ-Unsup, incDFM and DFM.}
\label{tab:CND_table} 
\vspace{-4mm}
\end{table*}

\vspace{-2mm}
\section{Experiments}
\vspace{-1mm}
We apply our uncertainty estimation method COUQ to a general open-world continual learning setting defined in \S\ref{sec:problem}. We demonstrate its utility and effectiveness on the following sub-tasks therein: (1) Continual Novelty Detection, results in \S\ref{sec:CND}; (2) Continual Active Learning, results in \S\ref{sec:CONL}. 


\subsection{Experimental Setup}
\label{sec:implementation}

\paragraph{Model:} The backbone $g$ used in our method and all baselines is a frozen, pre-trained deep model. This is a common practice in transfer learning in CL \cite{rios2020lifelong, rios2022incdfm}, and is theoretically based on the principle that low-level visual features obtained from a frozen model are thought to be task nonspecific and do not need to be constantly re-learned during CL tasks. We tested over 3 different pretrained foundation backbones: ResNet50 \cite{he2016deep} pre-trained on ImageNet1K via unsupervised SwAV \cite{caron2020unsupervised} and ViTs16 or ViTb16 \cite{alexey2020image} pre-trained on Imagenet1K via unsupervised DINO \cite{caron2021emerging}. 
The features $u$ are used for computation of the uncertainty scores in Eqs. (\ref{eq:uncertainty_scores}) and (\ref{eq:uncertainty-uncanny}) and also inputted to the classifier $f$ for eventual continual class-prediction. $f$ is implemented as a one hidden-layer perceptron (of size 4096).

As described in \S\ref{sec:steps-approach}, we test two variants of the novelty mapper $M^{i}(u)$ to assign class-ids to confidently novel samples:  (1) a K-means clustering in a fully unsupervised scenario that is trained to cluster the confident novel samples, and (2) a fully-connected layer (different from the main classifier $f$) on top of the frozen backbone in the semi-supervised scenario. Here, the mapper is trained with cross-entropy loss on the few actively labeled samples in addition to the confident novel samples pseudolabeled samples from the previous inner-loop iteration.
Finally, while COUQ is agnostic to which CL method is used to prevent catastrophic forgetting of the continual classifier, we showcase results employing a conventional CL technique termed ``Experience Replay'' (ER) \cite{rolnick2019experience, buzzega2021rethinking}. In ER methods, a limited number of exemplars from the old-classes must be stored in a buffer of fixed size $B$. These are used in conjunction with the novel samples to train the classifier without catastrophic forgetting. We set the $B$ to 5000 for Im21K-OOD and Places, and 2500 for Cifar100, Eurosat, and Plants. Further details about COUQ being used with ER are shared in the supplementary.

\vspace{-4mm}
\paragraph{Experiments:} Overall, we test Continual Novelty Detection and Continual Open-World Classification on 5 diverse datasets: Imagenet21K-OOD (Im21K-OOD) \cite{ridnik2021imagenet21k}, Places365-OOD (Places) \cite{zhou2017places}, Eurosat \cite{helber2019eurosat}, iNaturalist-Plants-20 (Plants) \cite{vanhorn2018inaturalist} and Cifar100-superclasses \cite{cifar100}. All of the aforementioned datasets were constructed to have class orthogonality (be out-of-distribution) with respect to Imagenet1K, which was used to pretrain the backbone (w/ exception of Cifar100). Further details on datasets are in the supplementary. At each incoming unlabeled pool (task), we fix a mixing ratio of 2:1 of old to new classes per task, with old classes drawn from a holdout set (0.35\% of each dataset prepared at experiment onset). We set pseudolabeling selection to $\alpha=20\%$ of samples predicted as novel, as described in \S\ref{sec:steps-approach} . After iterative updates of COUQ or baselines, for fair evaluation, we measure AUROC on an independent test set with the same ratio of old to new class samples as in the training pool (see \S\ref{sec:implementation}). For experiments not purposedly varying the continual class increments, we set default values as follows: 5 for Im21K-OODD, 5 for Places, 3 for both cifar100-superclasses and Plants, 2 for eurosat. More details can be found in the supplementary. 
\vspace{-4mm}
\paragraph{Baselines:} Since we test our approach in various settings, we  describe the relevant baselines chosen for each setting. 

\noindent\textit{(A) Baselines for only Continual Novelty Detection (CND):} (1) incDFM \cite{rios2022incdfm}, a method that includes an updatable continual novelty detector, but assumes single class novelties (see \S 2.2); (2) DFM \cite{ndiour2022subspace}, a precursor of incDFM originally proposed for non-continual novelty detection; 

\noindent\textit{(B) Baselines for only Continual Open-World Novelty Learning (CONL):} (3) CCIC \cite{Boschini2022} a semi-supervised CL model that adapts MixMatch \cite{berthelot2019mixmatch} to the CL setting; (4) GBCL, a few-shot continual active learning approach \cite{ayub2022few}; 

\noindent\textit{(C) Baselines used for both CND and CONL:} (5) (ER-variants) Semi-supervised baselines built upon the CL technique of Experience Replay \cite{rolnick2019experience,buzzega2021rethinking} based on \cite{vu2023active} for active continual learning. For these, uncertainty metrics such as (5.1) Entropy, (5.2) Margin, (5.3) Softmax are used to actively select the most uncertain samples for labeling which will be used in continual training updates. The same uncertainty metrics are then used to output scores for evaluation; (6) For further comparison, we introduce additional baselines built upon the previous item ER-variants but using the corresponding uncertainty quantification (e.g. Entropy, Margin or Softmax) to iteratively pseudolabel the most confident samples during the inner loop akin to our approach. 

\subsection{Continual Novelty Detection}
\label{sec:CND}
We first demonstrate the effectiveness of COUQ within the general problem of continual novelty detection (CND). Recall that at each task the novelty detector should be able to detect novelties reliably from an unknown and multiple number of novel classes. We evaluate CND performance using the common threshold-agnostic Area-under-receiver-operating-curve (AUROC) score. 
We show results for our method and all the relevant baselines over all 5 datasets and 3 different foundation model backbones in table \ref{tab:CND_table}. We test CND performance in both semi-supervised and unsupervised setups as described earlier. For the semi-supervised case, we fix AL budget at a default value of 1.25\% for all baselines (including our own method) and across all datasets, as detailed in Section \ref{sec:implementation}. Default class-increment -- the number of novel classes introduced at each task -- is indicated in parenthesis by the dataset name in the table. Entries in Table \ref{tab:CND_table} are averaged AUROC scores over all continual tasks. We plot CND performance (AUROC) over continual tasks (time) in Fig. \ref{fig:plots_CND} (left, A.1, B.1). Key takeways from Table \ref{tab:CND_table} are as follows: (1) Both our actively-supervised method version, COUQ (AL+P), and our unsupervised version, COUQ-Unsup (P), outperform competing methods by large margins over all experimental variations. In fact, our unsupervised variant overperforms even other methods that rely on semi-supervision. To note, semi-supervised methods are included in the first rows of the table, separated by a line. Naturally, COUQ-AL outperforms COUQ-Unsup. (2) Traditional uncertainty metrics --  Entropy, Margin, Softmax -- which are computed from the continual classification decision boundary perform poorly overall. One likely reason is the compounded negative effect of error-propagation (miss-identified samples) together with the pressures of catastrophic forgetting on the classification decision boundary. Note that in both ER- and PseudoER-variants, its corresponding uncertainty metric is used to select active samples with highest uncertainty and pseudolabels of high-confidence (in case of PseudoER). (3) Finally, note that PseudoER variants fail to consistently outperform ER. This is because, unlike our method, they are unable to produce high-quality, high-confidence pseudolabels. This highlights the importance of our COUQ uncertainty metric \ref{eq:uncertainty-uncanny} in measuring pseudolabel confidence. 

\vspace{-2mm}
\paragraph{Varying experimental parameters:} Next, we explore the impact of varying experimental parameters. First, we test with different AL budgets (from $0.625\%$ to $5\%$) for which the results are shown in Fig. \ref{fig:plots_CND} (Center, A.2-B.2). Our method -- both supervised and unsupervised -- continues to outperform baselines over a wide range of AL-budgets. Finally, we vary the class-increments from their default values with results plotted in Fig \ref{fig:plots_CND} (right, A.3 and B.3). We see that the compared SOTA novelty detector incDFM \cite{rios2022incdfm} performs quite well for the increment of one novel class per task, for which it was originally proposed. However, when the class increment increases, this method degrades in performance because it groups multiple novel classes together without distinction, which severely hurts detection capacity. We note a similar pattern in DFM. In comparison, due the clustering-based novelty mapper, our unsupervised COUQ-Unsup variant is able to significantly better model the novelty distribution along time and class increments.

\begin{table}[t]
\scriptsize
\centering
\addtolength{\tabcolsep}{-0.43em}
\begin{tabular}{c|c|c|c|c|c|c}
\toprule
{Ablations (R50)} & {Im21K} & {Eurosat} & {Plants} &{Cifar100} & {Places} & {Average (Change)}\\ 
  \midrule
  GT-Sup & 97.7 & 99 & 84.6 & 90.3 & 80.2& 90.4 ($\uparrow 2.2$) \\
  \midrule
  AL-Amb + P & \textbf{98} & \textbf{99.2} & \textbf{83.4} & \textbf{81.6} & \textbf{78.7} & \textbf{88.2} (0) \\
  AL-Top + P& 96.7 & 96.5 & 71.0 & 81.9 & 70.2&83.3 ($\downarrow 4.9$)\\
  AL-Rand +P &  97.7 & 99 & 75.4 & 81 & 78.6 & 86.3 ($\downarrow 2.9$)\\
  No-iters + P & 93.9 & 93.4 & 72.4 & 71.5 & 68.7&80.0 ($\downarrow 8.3$)\\
  AL-Amb & 92.6 & 91.8 & 63.8 & 74.5 & 64.6&77.5 ($\downarrow 10.7$)\\
  \hline
\end{tabular}
\vspace{-2mm}
\caption{AUROC results from ablations of COUQ on CND}
\label{tab:CND-ablations}
\vspace{-5mm}
\end{table}


\vspace{-3mm}
\paragraph{Ablations:} Next, we perform ablations to highlight the impact of various components of our proposed method. The results are presented in Table \ref{tab:CND-ablations}. In the first row (\textit{GT-Sup} or ground-truth supervision), \textit{all} the confident novel samples identified in a task are sent to a labeler with ground-truth label instead of a limited number of judiciously chosen ones. This is unrealistic in a real-world setting given high costs of labeling. Hence, it represents a conceptual upper-bound of performance, and there is no error-propagation between task transitions. 
For the remaining variants described next, an active budget of $1.25\%$ is assumed as before. These include: \textit{AL-Amb} - Querying samples ambiguous uncertainty scores as described in \S\ref{sec:COUQ-AL} for active labeling. This is the default strategy; (2) \textit{AL-Top} - Querying samples with highest uncertainty scores (i.e. most-confidently novel samples) for active labeling rather than ambiguous samples as in COUQ; (3) \textit{AL-Random} - Using a random selection of samples for active labeling; (4) \textit{No-Iters} - performing COUQ (with default stretgy) in oneshot rather than over multiple inner-loop iterations. In this case, we use all supervision budget upfront at $i=0$ and then also pseudolabel in one-shot prior to updating $S(i)$ \ref{eq:uncertainty-uncanny}. We additionally use `P' to denote when pseudolabeling is used in addition to active sample selection. Hence, the last row in Table \ref{tab:CND-ablations} contains results with AL only without pseudolabeling. First, we observe a small drop (2.2\%) in performance of our method relative to the fully-labeled \textit{GT-Sup} case. This indicates that for CND, COUQ manages to minimize the impact of error propagation. Next, we observe that other active labeling strategies \textit{AL-Top} or \textit{AL-Rand} decrease performance by 4.9\% and 2.9\% respectively, underscoring the informativeness of querying ambiguous samples for AL with the goal of continual novelty detection. We observe the importance of minimizing error propagation via our method's iterativeness since \textit{No-Iters} results in an significant 8.3\% decrease in performance. Finally, we show that pseudolabeling among the multiple novel classes detected is fundamental to performance given the AL budget's tiny size. Excluding pseudolabeling results in 10.7\% average decrease. 


\subsection{Continual Open-World Novelty Learning}
\label{sec:CONL}
Here we apply COUQ to Continual Open-World classification/learning (which we term CONL). The setting is identical to that described in \S\ref{sec:implementation}. However, in addition to novelty detection, the goal is also to learn to continuously classify and thus incorporate novel class samples into knowledge continuously. 
Within this general framework, we demonstrate two ways in which COUQ can be applied to help solve CONL. The first is using COUQ to decide which samples to actively label at each task. The second is to use it to reliably rank samples by their confidence of being of a given novel class $m$, from which it can be derived a reliable pseudolabeling algorithm. We finally combine the two aforementioned uses (Active and Pseudo) to offer a comprehensive and robust response to the CONL problem. 

\subsubsection{Active Labeling}

\label{sec:CONL-AL}
\begin{table}[h]
\scriptsize
\centering
\addtolength{\tabcolsep}{-0.43em}
\begin{tabular}{c|c|c|c|c|c|c}
\toprule
{Active Heuristic} & {Im21K} & {Eurosat} & {Plants} &{Cifar100} & {Places} & {Average}\\ 
  \midrule
  \midrule
  AL-Amb & \textbf{79.5} & \textbf{94.3}& \textbf{45.2}& \textbf{65.1} & 42.0 & \textbf{65.2}\\
  AL-Top & 76.9 & 82.7& 42.7& 56.8& 41.5 & 60.1\\
  No-Iters & 72.1 & 92.9& 41.9& 59.3& 38.3 & 60.9\\
  \midrule
  Entropy & 68.7 & 87.7 & 36.3& 54.2& 37.7 & 56.9\\
  Margin & 74.5 & 92.9& 43.1& 60.7& \textbf{42.5} & 62.7\\
  Max & 69.4& 88.7& 41.6& 58.0&37.3 & 59.0\\
  GBCL & 73.4& 93.2& 40& 60.2& 33.4 & 60.0\\
  Rand & 73.1& 92.5& 41.7 & 64.8 & 42.1 & 62.8\\
  \hline
\end{tabular}
\vspace{-2mm}
\caption{Effect of the AL strategy in CONL performance; Measured as average continual accuracy over all tasks and classes.}
\label{tab:CONL-AL}
\vspace{-3mm}
\end{table}

\begin{table*}[t]
\scriptsize
\centering
\addtolength{\tabcolsep}{-0.23em}
\begin{tabular}
{c|ccc|ccc|ccc|ccc|ccc|c} 
\toprule
{ER Methods} & \multicolumn{3}{c|}{Im21K} & \multicolumn{3}{c|}{Eurosat} & \multicolumn{3}{c|}{Plants} & \multicolumn{3}{c|}{Cifar100} & \multicolumn{3}{c}{Places} & {Average}\\ 
\midrule 
Method & R50 & ViT-B & ViT-S & R50 & ViT-B & ViT-S & R50 & ViT-B & ViT-S  & R50 & ViT-B & ViT-S & R50 & ViT-B & ViT-S & All\\ 
\midrule 
Oracle & 92.4& 83.2& 80.4& 98.0& 96.6& 94.8& 87.4& 74.9& 67.2& 76.5& 74.1& 70.2& 63.3& 50.6& 45.6 & 77.0\\ 
\midrule 
COUQ (AL+P) & \textbf{84.4} & \textbf{67.8}& \textbf{63.1}& \textbf{96.5}& \textbf{95.3}& \textbf{91.0}& \textbf{65.9}& \textbf{54.5}& \textbf{47.0}& \textbf{67.2}& \textbf{63.3}& \textbf{60.9}& \textbf{53.9}& \textbf{40.3}& \textbf{34.9} & \textbf{65.7}\\
GBCL (AL) & 73.4& 46.3& 43.1& 93.2& 80.1& 75.6& 40.0& 40.1& 32.8& 60.2& 48.8& 46.6& 33.4&30.5 &27.9 & 51.5\\  
CCIC (AL+S)& 58.0& -& -& 79.5& -& -& 34.7& -& -& 34.8& -& -& 26.5& -& - & 46.7\\  
Margin (AL)& 74.5& 46.1& 43.7& 92.9& 84.6& 77.8& 43.1& 30.9& 33.6& 60.7& 45.8& 44.5& 42.5& 27.6& 24.7 & 51.5\\ 
Margin (AL+P)& 70.1& 46.7& 44.9& 86.3& 90.1& 87.1& 49.7& 40.0& 38.8& 60.7& 54.4& 52.5& 42.6& 28.3& 28.2 & 54.7\\ 
\midrule 
\end{tabular}
\vspace{-2mm}
\caption{Continual Open-World Novelty Learning with Active Supervision ("AL") -"P" and "S" indicate pseudo-labeling and other semi-supervised technique respectively.}
\label{tab:CNL_table} 
\vspace{-5mm}
\end{table*}

We first isolate the effects of using COUQ for Active labeling only, with results in Table \ref{tab:CONL-AL}. We do so by removing pseudolabeling in Eq. \ref{eq:uncertainty-uncanny}, and using only the actively labeled samples for (i) updating the iterative metric, and (ii) updating the downstream continual classifier. The first three rows of Table \ref{tab:CONL-AL} show variants of AL using COUQ, same as those used in \S\ref{sec:CND}. Note however that `P' is omitted from their names since pseudolabeling is not used.
Similar to CND ablations, here too we see an improvement of \textit{AL-Amb} over \textit{AL-Top} and \textit{No-iters}.
Overall, COUQ with ambiguous selection strategy consistently outperforms other SOTA continual active learning baselines. The last row contains the lower-bound of random sample selection. 

\subsubsection{Pseudo Labeling}
\label{sec:CONL-P}
\begin{table}[h]
\scriptsize
\centering
\addtolength{\tabcolsep}{-0.47em}
\begin{tabular}{c|c|c|c|c|c|c}
\toprule
{Pseudolabeling (Rand-Sup)} & {Im21K} & {Eurosat} & {Plants} &{Cifar100} & {Places} & {Average}\\ 
  \midrule
  COUQ(P; Default) & 76.4 & \textbf{95.2} & \textbf{65.9} & \textbf{65.6} & \textbf{48.1}  & \textbf{70.2}\\
  COUQ(P; oneshot) & \textbf{78.9} & 93.9 & 43 & 63.7 & 43.7 & 64.6\\
  \midrule
  Entropy & 64.6 & 92.3 & 46.1 & 59.7 & 38.6 & 60.3\\
  Margin & 65 & 90.3 & 48 & 60.8 & 40.7 & 61.0\\
  Max & 62.1 & 89.9 & 50.5 & 60.5 & 40.4 & 60.7\\
  \midrule
  None & 73.1 & 92.5 & 41.7 & 64.8 & 42.1 & 62.8\\
  \hline
\end{tabular}
\vspace{-2mm}
\caption{Effect of Uncertainty scoring in Pseudolabeling in CONL; Measured as average CL accuracy over all tasks and classes.}
\label{tab:CONL-P}
\vspace{-3mm}
\end{table}

Next, we isolate the effect of COUQ when used only for pseudolabeling by adopting a random selection strategy for labeling instead. 
Table \ref{tab:CONL-P} shows that for pseudolabeling, COUQ overperforms conventional uncertainty metrics such as Margin. 
Note that for COUQ, the pseudolabels are used to update both Eq. (\ref{eq:uncertainty-uncanny}) and the downstream continual classifier. In the case of baselines, it is only the latter. 
Similar to previous results, removing the iterativeness (\textit{COUQ(P;oneshot}) leads to an 8\% decrease in performance. Most importantly, only COUQ pseudolabeling is consistently superior to abstaining from using pseudolabeling (lowerbound \textit{Rand}) and updating via only the few randomly labeled samples. Lastly, we want to emphasize that there are several semi-supervised learning techniques, which exploit the use of unlabeled data akin to pseudolabeling, and which are orthogonal to our method. Some examples are Consistency propagation, semi-supervised contrastive losses \cite{ma2024active,Boschini2022}, etc. These approaches can be used in tandem with COUQ and we leave that for future work. 
\subsubsection{Combining Active and pseudolabeling}
\label{sec:CONL-ALandP}
\vspace{-3mm}

\begin{figure}[h]
\centering
\includegraphics[width=8.2cm]{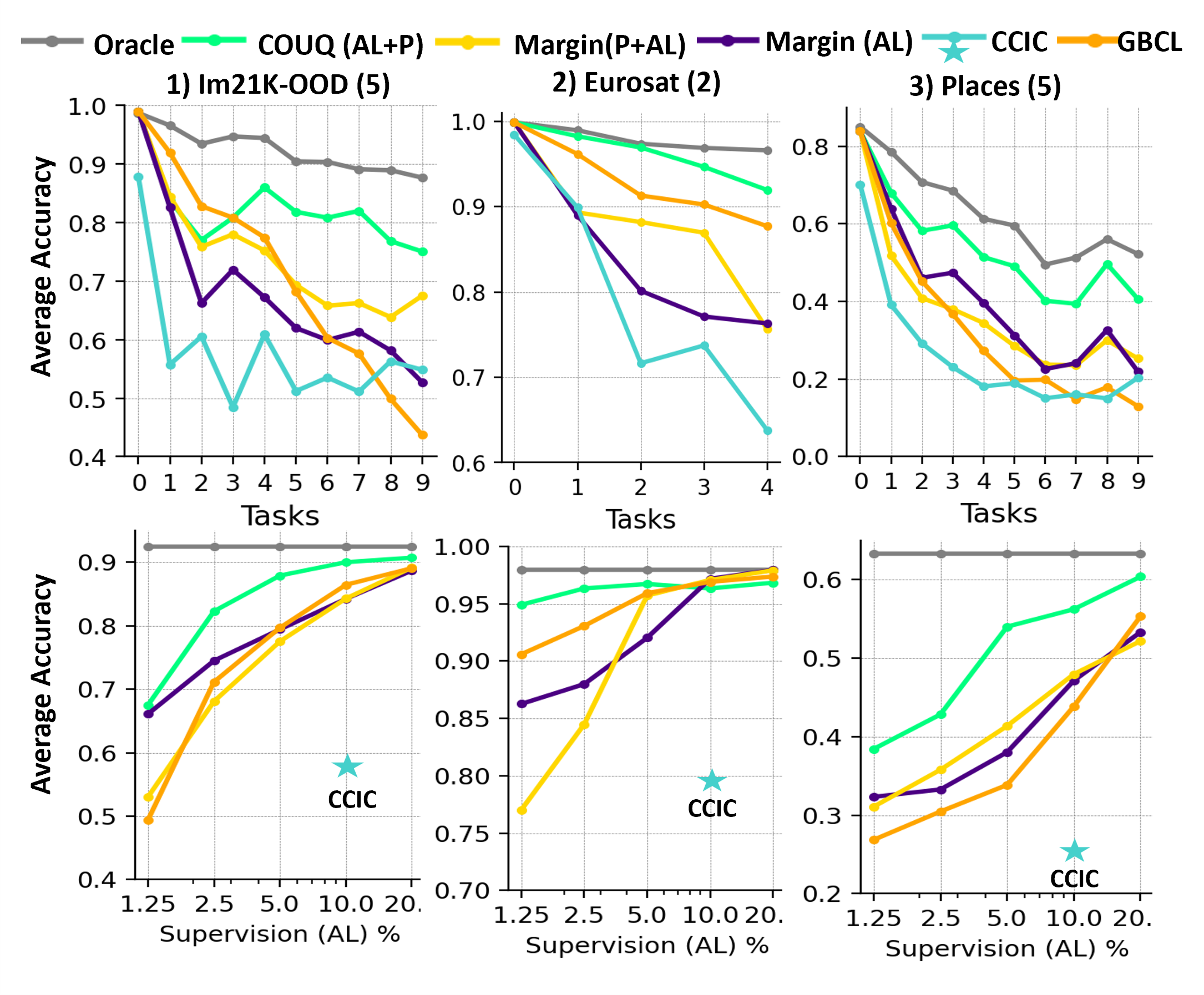}
\vspace{-4mm}
\caption{(Row 1) Continual classification accuracy over continual tasks during Continual Open-World Learning. The number of novel classes introduced per task for each dataset is in parenthesis. (Row 2) Results varying AL budget.}
\vspace{-2mm}
\label{fig:plots_CNL}
\end{figure}

Finally, we assess the performance of COUQ on a complete open-world CL pipeline comprising active labeling, pseudolabeling, and novelty detection.
Table \ref{tab:CNL_table} shows the cumulative average accuracy of the continual-learning classifier at the end of all tasks for all 5 datasets and 3 architecture variations. The ``Oracle'' method constitutes an upper-bound. It has perfect knowledge of old and new class labels (100\% supervision) and is trained using the same architecture and experience replay hyper-parameters as COUQ and all baselines. Overall, our method COUQ(AL+P), which includes both uncertainty-aware active and pseudolabeling via \ref{eq:uncertainty-uncanny}, outperforms all baselines by a large margin, even with stringent labeling budgets of $2.5-5\%$. When comparing COUQ(AL+P) to the ablated COUQ(AL) and COUQ(P) in sections \ref{sec:CONL-AL} and \ref{sec:CONL-P}, we see a clear boost in performance. By focusing AL strategy on including ambiguous novel samples and pseudolabeling on confident novel samples, COUQ(AL+P) is better qualified for the difficult problem of CONL. With respect to SOTA baselines, we see the classic Offline AL strategies adapted for ER (e.g.Margin, Entropty, Max) overall performs poorly. The main reason is that when datasets are more challenging and classes are presented in continual order, pressures from decreased accuracy and catastrophic forgetting diminish the trustworthiness of these AL metrics computed from the logit decision boundary. Moreover, as can be gauged by varying the AL budget, at low AL ratios, ER-AL variants decay abruptly. Additionally, baseline GBCL which was specifically developed for few-shot active open-world CL and is also trained solely from Actively labeled samples also under-performs. 
Fig. \ref{fig:plots_CNL} (row-2) analyzes the effect of varying the tiny active labeling budget. We show COUQ over-performs the other methods over a large interval of supervision budgets (all tested).  Fig. \ref{fig:plots_CNL} (row-1) shows cumulative continual accuracy results over all tasks. 
Note that CCIC (blue line or star symbol) drastically under-performs all other approaches even with a large supervision budget of 10\%. As most semi-supervised CL methods, it originally assumed old and new classes would not co-occur and cannot properly quantify uncertainty when this assumption is lifted. Equivalent plots for other datasets are in the supplementary. 
    
\vspace{-3mm}
\section{Conclusion}
\vspace{-2mm}
We present an uncertainty quantification method specifically designed for continual open-world learning. With our method, we are able to assign high-confidence pseudo-labels which are reliable and that we show can significantly reduce labeling costs. Furthermore, also with our uncertainty method, we demonstrate that active querying based on novelty ambiguity may be more informative than merely selecting the most-likely novel samples. Our approach outperforms baselines across multiple datasets and experiments. 
Yet, several challenges remain, which we aim to address in future work. An example is how to continually query and update when distribution shifts of past classes (e.g., noise, illumination changes) co-occur with novel classes.

{
    \small
    \bibliographystyle{ieeenat_fullname}
    \bibliography{main}

\begin{thebibliography}{53}
\providecommand{\natexlab}[1]{#1}
\providecommand{\url}[1]{\texttt{#1}}
\expandafter\ifx\csname urlstyle\endcsname\relax
  \providecommand{\doi}[1]{doi: #1}\else
  \providecommand{\doi}{doi: \begingroup \urlstyle{rm}\Url}\fi

\bibitem[Aguilar et~al.(2023)Aguilar, Raducanu, Radeva, and Van~de Weijer]{Aguilar_2023_ICCV}
Eduardo Aguilar, Bogdan Raducanu, Petia Radeva, and Joost Van~de Weijer.
\newblock Continual evidential deep learning for out-of-distribution detection.
\newblock In \emph{Proceedings of the IEEE/CVF International Conference on Computer Vision (ICCV) Workshops}, pages 3444--3454, 2023.

\bibitem[Ahuja et~al.(2019)Ahuja, Ndiour, Kalyanpur, and Tickoo]{ahuja2019bdl_dfm}
Nilesh~A. Ahuja, Ibrahima~J. Ndiour, Trushant Kalyanpur, and Omesh Tickoo.
\newblock Probabilistic modeling of deep features for out-of-distribution and adversarial detection.
\newblock In \emph{Bayesian Deep Learning workshop, NeurIPS}, 2019.

\bibitem[Alexey(2020)]{alexey2020image}
Dosovitskiy Alexey.
\newblock An image is worth 16x16 words: Transformers for image recognition at scale.
\newblock \emph{arXiv preprint arXiv: 2010.11929}, 2020.

\bibitem[Aljundi et~al.(2022)Aljundi, Reino, Chumerin, and Turner]{aljundi2022continual}
Rahaf Aljundi, Daniel~Olmeda Reino, Nikolay Chumerin, and Richard~E Turner.
\newblock Continual novelty detection.
\newblock In \emph{Conference on Lifelong Learning Agents}, pages 1004--1025. PMLR, 2022.

\bibitem[Ayub and Fendley(2022)]{ayub2022few}
Ali Ayub and Carter Fendley.
\newblock Few-shot continual active learning by a robot.
\newblock \emph{Advances in Neural Information Processing Systems}, 35:\penalty0 30612--30624, 2022.

\bibitem[Ayub and Fendley(2024)]{GBCL}
Ali Ayub and Carter Fendley.
\newblock Few-shot continual active learning by a robot.
\newblock In \emph{Proceedings of the 36th International Conference on Neural Information Processing Systems}. Curran Associates Inc., 2024.

\bibitem[Bagus et~al.(2022)Bagus, Gepperth, and Lesort]{bagus2022supervised}
Benedikt Bagus, Alexander Gepperth, and Timothée Lesort.
\newblock Beyond supervised continual learning: a review.
\newblock \emph{arXiv preprint arXiv:2208.14307}, 2022.

\bibitem[Berthelot et~al.(2019)Berthelot, Carlini, Goodfellow, Papernot, Oliver, and Raffel]{berthelot2019mixmatch}
David Berthelot, Nicholas Carlini, Ian Goodfellow, Nicolas Papernot, Avital Oliver, and Colin~A Raffel.
\newblock Mixmatch: A holistic approach to semi-supervised learning.
\newblock \emph{Advances in neural information processing systems}, 32, 2019.

\bibitem[Boschini et~al.(2022{\natexlab{a}})Boschini, Buzzega, Bonicelli, Porrello, and Calderara]{Boschini2022}
Matteo Boschini, Pietro Buzzega, Lorenzo Bonicelli, Angelo Porrello, and Simone Calderara.
\newblock Continual semi-supervised learning through contrastive interpolation consistency.
\newblock \emph{Pattern Recognition Letters}, 162:\penalty0 9–14, 2022{\natexlab{a}}.

\bibitem[Boschini et~al.(2022{\natexlab{b}})Boschini, Buzzega, Bonicelli, Porrello, and Calderara]{Boschini_2022}
Matteo Boschini, Pietro Buzzega, Lorenzo Bonicelli, Angelo Porrello, and Simone Calderara.
\newblock Continual semi-supervised learning through contrastive interpolation consistency.
\newblock \emph{Pattern Recognition Letters}, 162:\penalty0 9–14, 2022{\natexlab{b}}.

\bibitem[Brinker(2003)]{brinker}
Klaus Brinker.
\newblock Incorporating diversity in active learning with support vector machines.
\newblock In \emph{International Conference on Machine Learning}, 2003.

\bibitem[Buzzega et~al.(2021)Buzzega, Boschini, Porrello, and Calderara]{buzzega2021rethinking}
Pietro Buzzega, Matteo Boschini, Angelo Porrello, and Simone Calderara.
\newblock Rethinking experience replay: a bag of tricks for continual learning.
\newblock In \emph{2020 25th International Conference on Pattern Recognition (ICPR)}, pages 2180--2187. IEEE, 2021.

\bibitem[Caron et~al.(2020)Caron, Misra, Mairal, Goyal, Bojanowski, and Joulin]{caron2020unsupervised}
Mathilde Caron, Ishan Misra, Julien Mairal, Priya Goyal, Piotr Bojanowski, and Armand Joulin.
\newblock Unsupervised learning of visual features by contrasting cluster assignments.
\newblock \emph{Advances in Neural Information Processing Systems}, 33:\penalty0 9912--9924, 2020.

\bibitem[Caron et~al.(2021)Caron, Touvron, Misra, J{\'e}gou, Mairal, Bojanowski, and Joulin]{caron2021emerging}
Mathilde Caron, Hugo Touvron, Ishan Misra, Herv{\'e} J{\'e}gou, Julien Mairal, Piotr Bojanowski, and Armand Joulin.
\newblock Emerging properties in self-supervised vision transformers.
\newblock In \emph{Proceedings of the IEEE/CVF international conference on computer vision}, pages 9650--9660, 2021.

\bibitem[Cheung et~al.(2019)Cheung, Terekhov, Chen, Agrawal, and Olshausen]{cheung2019superposition}
Brian Cheung, Alexander Terekhov, Yubei Chen, Pulkit Agrawal, and Bruno Olshausen.
\newblock Superposition of many models into one.
\newblock \emph{Advances in neural information processing systems}, 32, 2019.

\bibitem[Fei and Liu(2016)]{fei2016breaking}
Geli Fei and Bing Liu.
\newblock Breaking the closed world assumption in text classification.
\newblock In \emph{Proceedings of the 2016 Conference of the North American Chapter of the Association for Computational Linguistics: Human Language Technologies}, pages 506--514, 2016.

\bibitem[French(1999)]{french1999catastrophic}
Robert~M French.
\newblock Catastrophic forgetting in connectionist networks.
\newblock \emph{Trends in cognitive sciences}, 3\penalty0 (4):\penalty0 128--135, 1999.

\bibitem[Gal and Ghahramani(2016)]{gal2016dropout}
Yarin Gal and Zoubin Ghahramani.
\newblock Dropout as a bayesian approximation: Representing model uncertainty in deep learning.
\newblock In \emph{international conference on machine learning}, pages 1050--1059, 2016.

\bibitem[Gal et~al.(2017)Gal, Islam, and Ghahramani]{gal2017deep}
Yarin Gal, Riashat Islam, and Zoubin Ghahramani.
\newblock Deep bayesian active learning with image data.
\newblock In \emph{International Conference on Machine Learning}, pages 1183--1192. PMLR, 2017.

\bibitem[Han et~al.(2019)Han, Vedaldi, and Zisserman]{han2019learning}
Kai Han, Andrea Vedaldi, and Andrew Zisserman.
\newblock Learning to discover novel visual categories via deep transfer clustering.
\newblock In \emph{Proceedings of the IEEE/CVF International Conference on Computer Vision}, pages 8401--8409, 2019.

\bibitem[He et~al.(2016)He, Zhang, Ren, and Sun]{he2016deep}
Kaiming He, Xiangyu Zhang, Shaoqing Ren, and Jian Sun.
\newblock Deep residual learning for image recognition.
\newblock In \emph{Proceedings of the IEEE conference on computer vision and pattern recognition}, pages 770--778, 2016.

\bibitem[Helber et~al.(2019)Helber, Bischke, Dengel, and Borth]{helber2019eurosat}
Patrick Helber, Benjamin Bischke, Andreas Dengel, and Damian Borth.
\newblock Eurosat: A novel dataset and deep learning benchmark for land use and land cover classification, 2019.

\bibitem[Hendrycks and Gimpel(2017)]{hendrycks2016baseline}
Dan Hendrycks and Kevin Gimpel.
\newblock A baseline for detecting misclassified and out-of-distribution examples in neural networks.
\newblock 2017.

\bibitem[Horn et~al.(2018)Horn, Aodha, Song, Cui, Sun, Shepard, Adam, Perona, and Belongie]{vanhorn2018inaturalist}
Grant~Van Horn, Oisin~Mac Aodha, Yang Song, Yin Cui, Chen Sun, Alex Shepard, Hartwig Adam, Pietro Perona, and Serge Belongie.
\newblock The inaturalist species classification and detection dataset, 2018.

\bibitem[Kang et~al.(2023)Kang, Fini, Nabi, Ricci, and Alahari]{ordisco}
Zhiqi Kang, Enrico Fini, Moin Nabi, Elisa Ricci, and Karteek Alahari.
\newblock A soft nearest-neighbor framework for continual semi-supervised learning.
\newblock In \emph{Proceedings of the IEEE/CVF International Conference on Computer Vision}, pages 11868--11877, 2023.

\bibitem[Kirkpatrick et~al.(2017)Kirkpatrick, Pascanu, Rabinowitz, Veness, Desjardins, Rusu, Milan, Quan, Ramalho, Grabska-Barwinska, et~al.]{kirkpatrick2017overcoming}
James Kirkpatrick, Razvan Pascanu, Neil Rabinowitz, Joel Veness, Guillaume Desjardins, Andrei~A Rusu, Kieran Milan, John Quan, Tiago Ramalho, Agnieszka Grabska-Barwinska, et~al.
\newblock Overcoming catastrophic forgetting in neural networks.
\newblock \emph{Proceedings of the national academy of sciences}, 114\penalty0 (13):\penalty0 3521--3526, 2017.

\bibitem[Krizhevsky(2012)]{cifar100}
Alex Krizhevsky.
\newblock Learning multiple layers of features from tiny images.
\newblock \emph{University of Toronto}, 2012.

\bibitem[Lee et~al.(2018)Lee, Lee, Lee, and Shin]{lee2018simple}
Kimin Lee, Kibok Lee, Honglak Lee, and Jinwoo Shin.
\newblock A simple unified framework for detecting out-of-distribution samples and adversarial attacks.
\newblock In \emph{Advances in Neural Information Processing Systems}, pages 7167--7177, 2018.

\bibitem[Lewis and Gale(1994)]{lewis_gale}
David~D. Lewis and William~A. Gale.
\newblock A sequential algorithm for training text classifiers.
\newblock In \emph{SIGIR '94}, pages 3--12, London, 1994. Springer London.

\bibitem[Liang et~al.(2018)Liang, Li, and Srikant]{liang2018enhancing}
Shiyu Liang, Yixuan Li, and R Srikant.
\newblock Enhancing the reliability of out-of-distribution image detection in neural networks.
\newblock 2018.

\bibitem[Ma et~al.(2024)Ma, Zhu, Zhong, Zhang, and Liu]{ma2024active}
Shijie Ma, Fei Zhu, Zhun Zhong, Xu-Yao Zhang, and Cheng-Lin Liu.
\newblock Active generalized category discovery.
\newblock In \emph{Proceedings of the IEEE/CVF Conference on Computer Vision and Pattern Recognition}, pages 16890--16900, 2024.

\bibitem[Ndiour et~al.(2020)Ndiour, Ahuja, and Tickoo]{ndiour2020probabilistic}
Ibrahima Ndiour, Nilesh~A Ahuja, and Omesh Tickoo.
\newblock Out-of-distribution detection with subspace techniques and probabilistic modeling of features.
\newblock \emph{arXiv preprint arXiv:2012.04250}, 2020.

\bibitem[Ndiour et~al.(2022)Ndiour, Ahuja, and Tickoo]{ndiour2022subspace}
Ibrahima~J Ndiour, Nilesh~A Ahuja, and Omesh Tickoo.
\newblock Subspace modeling for fast out-of-distribution and anomaly detection.
\newblock In \emph{2022 IEEE International Conference on Image Processing (ICIP)}, pages 3041--3045. IEEE, 2022.

\bibitem[Parisi et~al.(2019)Parisi, Kemker, Part, Kanan, and Wermter]{parisi2019continual}
German~I Parisi, Ronald Kemker, Jose~L Part, Christopher Kanan, and Stefan Wermter.
\newblock Continual lifelong learning with neural networks: A review.
\newblock \emph{Neural Networks}, 113:\penalty0 54--71, 2019.

\bibitem[Parmar et~al.(2023)Parmar, Chouhan, Raychoudhury, and Rathore]{parmar2023open}
Jitendra Parmar, Satyendra Chouhan, Vaskar Raychoudhury, and Santosh Rathore.
\newblock Open-world machine learning: applications, challenges, and opportunities.
\newblock \emph{ACM Computing Surveys}, 55\penalty0 (10):\penalty0 1--37, 2023.

\bibitem[Rebuffi et~al.(2017)Rebuffi, Kolesnikov, Sperl, and Lampert]{rebuffi2017icarl}
Sylvestre-Alvise Rebuffi, Alexander Kolesnikov, Georg Sperl, and Christoph~H Lampert.
\newblock icarl: Incremental classifier and representation learning.
\newblock In \emph{Proceedings of the IEEE conference on Computer Vision and Pattern Recognition}, pages 2001--2010, 2017.

\bibitem[Ren et~al.(2019)Ren, Liu, Fertig, Snoek, Poplin, Depristo, Dillon, and Lakshminarayanan]{ren2019likelihood}
Jie Ren, Peter~J Liu, Emily Fertig, Jasper Snoek, Ryan Poplin, Mark Depristo, Joshua Dillon, and Balaji Lakshminarayanan.
\newblock Likelihood ratios for out-of-distribution detection.
\newblock In \emph{Advances in Neural Information Processing Systems}, pages 14707--14718, 2019.

\bibitem[Ren et~al.(2021)Ren, Xiao, Chang, Huang, Li, Gupta, Chen, and Wang]{ren_survey_2021}
Pengzhen Ren, Yun Xiao, Xiaojun Chang, Po-Yao Huang, Zhihui Li, Brij~B. Gupta, Xiaojiang Chen, and Xin Wang.
\newblock A {Survey} of {Deep} {Active} {Learning}.
\newblock \emph{ACM Computing Surveys}, 54\penalty0 (9):\penalty0 180:1--180:40, 2021.

\bibitem[Ridnik et~al.(2021)Ridnik, Ben-Baruch, Noy, and Zelnik-Manor]{ridnik2021imagenet21k}
Tal Ridnik, Emanuel Ben-Baruch, Asaf Noy, and Lihi Zelnik-Manor.
\newblock Imagenet-21k pretraining for the masses, 2021.

\bibitem[Rios and Itti(2018)]{rios2018closed}
Amanda Rios and Laurent Itti.
\newblock Closed-loop memory gan for continual learning.
\newblock \emph{arXiv preprint arXiv:1811.01146}, 2018.

\bibitem[Rios and Itti(2020)]{rios2020lifelong}
Amanda Rios and Laurent Itti.
\newblock Lifelong learning without a task oracle.
\newblock In \emph{2020 IEEE 32nd International Conference on Tools with Artificial Intelligence (ICTAI)}, pages 255--263. IEEE, 2020.

\bibitem[Rios et~al.(2022)Rios, Ahuja, Ndiour, Genc, Itti, and Tickoo]{rios2022incdfm}
Amanda Rios, Nilesh Ahuja, Ibrahima Ndiour, Utku Genc, Laurent Itti, and Omesh Tickoo.
\newblock incdfm: Incremental deep feature modeling for continual novelty detection.
\newblock In \emph{European Conference on Computer Vision}, pages 588--604. Springer, 2022.

\bibitem[Rolnick et~al.(2019)Rolnick, Ahuja, Schwarz, Lillicrap, and Wayne]{rolnick2019experience}
David Rolnick, Arun Ahuja, Jonathan Schwarz, Timothy Lillicrap, and Gregory Wayne.
\newblock Experience replay for continual learning.
\newblock \emph{Advances in Neural Information Processing Systems}, 32, 2019.

\bibitem[Sener and Savarese(2018)]{sener2018active}
Ozan Sener and Silvio Savarese.
\newblock Active learning for convolutional neural networks: A core-set approach.
\newblock In \emph{International Conference on Learning Representations}, 2018.

\bibitem[Settles(2009)]{settles2009active}
Burr Settles.
\newblock Active learning literature survey.
\newblock Computer Sciences Technical Report 1648, University of Wisconsin--Madison, 2009.

\bibitem[Shu et~al.(2017)Shu, Xu, and Liu]{shu2017doc}
Lei Shu, Hu Xu, and Bing Liu.
\newblock Doc: Deep open classification of text documents.
\newblock In \emph{Proceedings of the 2017 Conference on Empirical Methods in Natural Language Processing}, pages 2911--2916, 2017.

\bibitem[Tong and Koller(2002)]{tong_koller}
Simon Tong and Daphne Koller.
\newblock Support vector machine active learning with applications to text classification.
\newblock \emph{J. Mach. Learn. Res.}, 2:\penalty0 45--66, 2002.

\bibitem[Vaze et~al.(2022)Vaze, Han, Vedaldi, and Zisserman]{vaze2022gcd}
Sagar Vaze, Kai Han, Andrea Vedaldi, and Andrew Zisserman.
\newblock Generalized category discovery.
\newblock In \emph{IEEE Conference on Computer Vision and Pattern Recognition}, 2022.

\bibitem[Vu et~al.(2023)Vu, Khadivi, Ghorbanali, Phung, and Haffari]{vu2023active}
Thuy-Trang Vu, Shahram Khadivi, Mahsa Ghorbanali, Dinh Phung, and Gholamreza Haffari.
\newblock Active continual learning: On balancing knowledge retention and learnability.
\newblock \emph{arXiv preprint arXiv:2305.03923}, 2023.

\bibitem[Wang et~al.(2022)Wang, Li, Feng, and Zhang]{haoqi2022vim}
Haoqi Wang, Zhizhong Li, Litong Feng, and Wayne Zhang.
\newblock Vim: Out-of-distribution with virtual-logit matching.
\newblock In \emph{Proceedings of the IEEE/CVF Conference on Computer Vision and Pattern Recognition}, 2022.

\bibitem[Wen et~al.(2021)Wen, Rios, Ge, and Itti]{wen2021beneficial}
Shixian Wen, Amanda Rios, Yunhao Ge, and Laurent Itti.
\newblock Beneficial perturbation network for designing general adaptive artificial intelligence systems.
\newblock \emph{IEEE Transactions on Neural Networks and Learning Systems}, 2021.

\bibitem[Yoo and Kweon(2019)]{yoo2019learning}
Donggeun Yoo and In~So Kweon.
\newblock Learning loss for active learning.
\newblock In \emph{Proceedings of the IEEE/CVF Conference on Computer Vision and Pattern Recognition}, pages 93--102, 2019.

\bibitem[Zhou et~al.(2017)Zhou, Lapedriza, Khosla, Oliva, and Torralba]{zhou2017places}
Bolei Zhou, Agata Lapedriza, Aditya Khosla, Aude Oliva, and Antonio Torralba.
\newblock Places: A 10 million image database for scene recognition.
\newblock \emph{IEEE Transactions on Pattern Analysis and Machine Intelligence}, 2017.

\end{thebibliography}
}

\end{document}